%% file: uai2025-main.tex
\documentclass[accepted]{uai2025} 
                        
\usepackage{algorithm}
\usepackage{algpseudocode}
\usepackage{subcaption}
\usepackage[american]{babel}
\usepackage{amssymb}
\usepackage{natbib} 
\usepackage{mathtools} 
\usepackage{booktabs} 
\usepackage{tikz} 

\usepackage{xargs}                      

\usepackage[colorinlistoftodos,prependcaption,textsize=tiny]{todonotes}
\newcommandx{\unsure}[2][1=]{\todo[linecolor=red,backgroundcolor=red!25,bordercolor=red,#1]{#2}}
\newcommandx{\change}[2][1=]{\todo[linecolor=blue,backgroundcolor=blue!25,bordercolor=blue,#1]{#2}}
\newcommandx{\info}[2][1=]{\todo[linecolor=OliveGreen,backgroundcolor=OliveGreen!25,bordercolor=OliveGreen,#1]{#2}}
\newcommandx{\improvement}[2][1=]{\todo[linecolor=Plum,backgroundcolor=Plum!25,bordercolor=Plum,#1]{#2}}
\newcommandx{\thiswillnotshow}[2][1=]{\todo[disable,#1]{#2}}

\title{Learning to Stop Overthinking at Test Time}

\author[1]{\href{mailto:hieubkls98@gmail.com}{Hieu Tran Bao}{}}
\author[2]{Nguyen Cong Dat}
\author[3]{Nguyen Duc Anh}
\author[4]{Hoang Thanh-Tung}
\affil[1]{%
    FPT IS AI R\&D Center, Ha Noi, Viet Nam
}
\affil[2]{%
    FPT Software AI Center, Ha Noi, Viet Nam
}
\affil[3]{%
    Kaopiz Solutions, Ha Noi, Viet Nam
  }

\affil[4]{%
    VNU - University of Engineering and Technology, Ha Noi, Viet Nam
  }

\begin{document}
\maketitle

\input{section/abstract}
\input{section/introduction}
\input{section/related_work}
\input{section/methodology}

\input{section/experiment}
\input{section/conclusion}


\input{output.bbl}
\newpage
\input{section/supplementary}

\end{document}

%% file: section/abstract.tex
\begin{abstract}  
Test time scaling is currently one of the most active research areas that shows promise after training time scaling has reached its limits.
Deep-thinking (DT) models are a class of recurrent models that can perform easy-to-hard generalization by assigning more compute to harder test samples.
However, due to their inability to determine the complexity of a test sample, DT models have to use a large amount of computation for both easy and hard test samples.
Excessive test time computation is wasteful and can cause the ``overthinking'' problem where more test time computation leads to worse results.
In this paper, we introduce a test time training method for determining the optimal amount of computation needed for each sample during test time.
We also propose Conv-LiGRU, a novel recurrent architecture for efficient and robust visual reasoning. 
Extensive experiments demonstrate that Conv-LiGRU is more stable than DT, effectively mitigates the ``overthinking'' phenomenon, and achieves superior accuracy.
\end{abstract}  

%% file: section/introduction.tex
\section{Introduction}
\label{sec:intro}

Recurrent Neural Networks (RNNs) have proven to be highly effective in tackling machine reasoning tasks, demonstrating remarkable capability to manage problems of different complexity levels within a single task \citep{orvieto2023resurrecting,de2024griffin,beck2024xlstm}.
However, traditional RNNs struggle to autonomously generalize to more complex problems beyond those encountered during training. 
RNN's sequential architecture and limited memory capacity hinder parallel training and scalability, causing them to fall behind Transformers.

Despite extensive research on the reasoning capabilities of recurrent models, they are mainly used for simple sequence processing tasks like prefix sum or sequence copying. 
Some studies \citep{eyzaguirre2020differentiable,veerabadran2023Adaptive} explored their use in visual reasoning but focus on time-dependent tasks like maze-solving or chess. 
Static environments without explicit reasoning steps like object recognition, remain underexplored.
Recent studies in vision-language reasoning have sought to integrate visual understanding into large language models (LLMs) \citep{lin2024vila,wang2024qwen2}. 
To enhance reasoning capability, these models employ Chain-of-Thought (CoT), generating step-by-step demonstrations of images, similar to methods used in LLMs \citep{dong2024insight,thawakar2025llamav}. 
However, such approaches often overlook the models' robustness to low-quality and noisy images.
Training solely on curated datasets makes them highly vulnerable to inappropriate prompts. 
Furthermore, the discrete nature of language-based and tree-based reasoning models could lead to exponential complexity as the models try to imitate discrete search algorithms like DFS, BFS, and A$^*$ \citep{lehnert2024beyond, yao2023tree}
In contrast, studies on latent language models have revealed promising signs of enhanced robustness \citep{hao2024training}, and the efficacy of inference in latent space has been confirmed in other fields \citep{rombach2022high,radford2021learning,videoworldsimulators2024}.
\todo[disable]{More on: Why is this an important problem? 
What are the new challenges this problem introduces? 
Why is it beneficial to study this problem in parallel with language reasoning? 
Evidence of humans doing both latent and language reasoning. The advantages of latent reasoning. 
Can robust latent OOD generalization benefit reasoning/systematic generalization?}

In this study, we take the first steps toward exploring the reasoning capabilities of visual recurrent models in latent space, in parallel with CoT techniques in LLMs. 
Motivated by Deep Thinking \citep{schwarzschild2021can,bansal2022endtoend}, our proposed model can generalize to tackle more complex problems at test time simply by iterating its recurrent units more times and no additional training is needed. 
Our approach enables zero-shot extrapolation to more challenging environments within the same task.
The ability to handle problems under various conditions enables the development of robust and adaptable models, which are crucial for real-world applications and have the potential to apply to LLMs. 
We explore the effectiveness of RNNs in object recognition tasks using the CIFAR10-C and CIFAR100-C \citep{cifarC} datasets.

The key contributions of this study are:
\begin{itemize}
    \item We explore the extrapolation capabilities of recurrent model architectures for simple visual reasoning tasks, specifically object recognition. 
    We demonstrate that recurrent models enable strong extrapolation while utilizing significantly fewer parameters compared to conventional feedforward networks.
    \item We show that the early stopping heuristic in previous works—which forces the model to halt as soon as possible—limits the extrapolation capabilities of RNNs. 
    To enhance performance, we propose using a self-supervised task to estimate the accuracy trend across iterations, allowing us to determine the optimal number of iterations for the main task.
    \item We propose Conv-LiGRU, a novel recurrent model for effective and compute-optimal visual reasoning. 
    \item Extensive analysis and experiments show that Conv-LiGRU is more stable than Conv-GRU, better mitigates the "overthinking" phenomenon, and achieves superior accuracy compared to previous methods.

\end{itemize}

%% file: section/related_work.tex
\section{Related Works}
\label{sec:relatedwork}

\textbf{Thinking in Language vs. Latent Space:} 
Recent studies, most notably OpenAI-O1 \cite{jaech2024openai}, have demonstrated that large language models (LLMs) can handle more complex tasks by thinking longer before answering in natural language.  
While it brings large language models closer to human-like thinking, this approach activates all layers at any time. 
Furthermore, its performance depends largely on the thought sequence's length, making it challenging to perform more difficult tasks with fewer computations.
Recently, Deepseek-R1 \cite{guo2025deepseek} has utilized a fraction of the large model for inference, allowing adaptation to edge devices. 
In this study, we focus on implicit reasoning embedded within recurrent models. 
Reasoning in the latent space allows models to synthesize and refine instance interpretability, akin to a conventional deep model.
\citet{xu2024lars} extends the latent reasoning to In-Context Learning, showing greater robustness with a faster inference time. 
As opposed to CoT reasoning, \citet{hao2024training} demonstrates that the continuous space can represent multiple alternative reasoning steps, thereby significantly expanding the model's search space.


\textbf{Test-time Training (TTT):} 
Another line of work coinciding with ours is learning at test time by directly updating model parameters on test data without supervision. 
Previous work has shown that TTT is robust to distribution shifts \citep{sun2020test,gandelsman2022test}, while \citet{muennighoff2025s1simpletesttimescaling} shows that a simple test-time training can beat OpenAI-O1 on the math question.
The connection between the RNN update mechanisms, the attention mechanisms, and the TTT has been highlighted in \citet{sun2024learning}, therefore a new TTT layer is proposed for the generation of long sequences. 
Our study leverages self-supervised tasks to estimate the optimal reasoning depth, improving the model's computational efficiency and reasoning performance.

\textbf{Deep Thinking:} 
\citet{schwarzschild2021can} demonstrated that recurrent models trained on simple tasks can generalize to harder ones simply by repeating a set of layers more times during testing. 
However, \citet{bansal2022endtoend} identified the "overthinking" problem, where longer inference time leads to worse performance.
To mitigate this issue, they proposed Recall architecture and Progressive Loss. 
The "Recall" architecture adds residual connections from the original input to every recurrent input, effectively preventing ``overthinking'' due to vanishing gradient.
Progressive Loss forces the outputs across iterations to be consistent, allowing the hidden representations to converge to a fixed point after \textit{some iterations}. 
\todo[disable]{should be more on the point: DT+RC models prevent overthinking with residual connections and forcing the hidden states to converge to fixed points}
\todo[disable]{Fixed??}
\citet{bansal2022endtoend} used a very large number of steps for all difficulty levels to ensure that the representations converged because they could not determine the optimal number of steps.
To address this, \citet{veerabadran2023Adaptive,Ballas2016Iclr} applied ACT \cite{Graves2016AdaptiveComutationTime}, using a sigmoidal unit to decide when to stop iterating. 
Since stopping probabilities cannot be predetermined, ACT added a "ponder cost" to encourage early termination. 
Building on this, \citet{PonderNet} restricted the stopping probabilities to a predefined prior, allowing control over the termination point by adjusting the prior.

\citet{schwarzschild2021can, bansal2022endtoend,veerabadran2023Adaptive} conducted experiments on logical tasks such as prefix sum, maze solving, chess, and pathfinding, demonstrating the logical extrapolation capabilities of recurrent models. 
However, the extrapolation ability of recurrent models has yet to be explored in classical computer vision tasks like object recognition.
\citet{cifarC} developed corruption datasets CIFAR10-C, CIFAR100-C, and ImageNetC corruption datasets to generalize the model by introducing 15 types of corruption at five different severity levels.
Noting the similarity between the experimental setup and the logical tasks, we show that recurrent models trained on lower noise level data can achieve robustness on harder noise levels. 
We further demonstrate that recurrent models can exhibit extrapolation capabilities in object recognition tasks, mirroring human perception. Under ideal, noise-free conditions, recognition is swift and effortless, whereas challenging conditions demand additional time for pattern identification.

\begin{figure}[!t]
    \centering
    \includegraphics[width=0.98\linewidth]{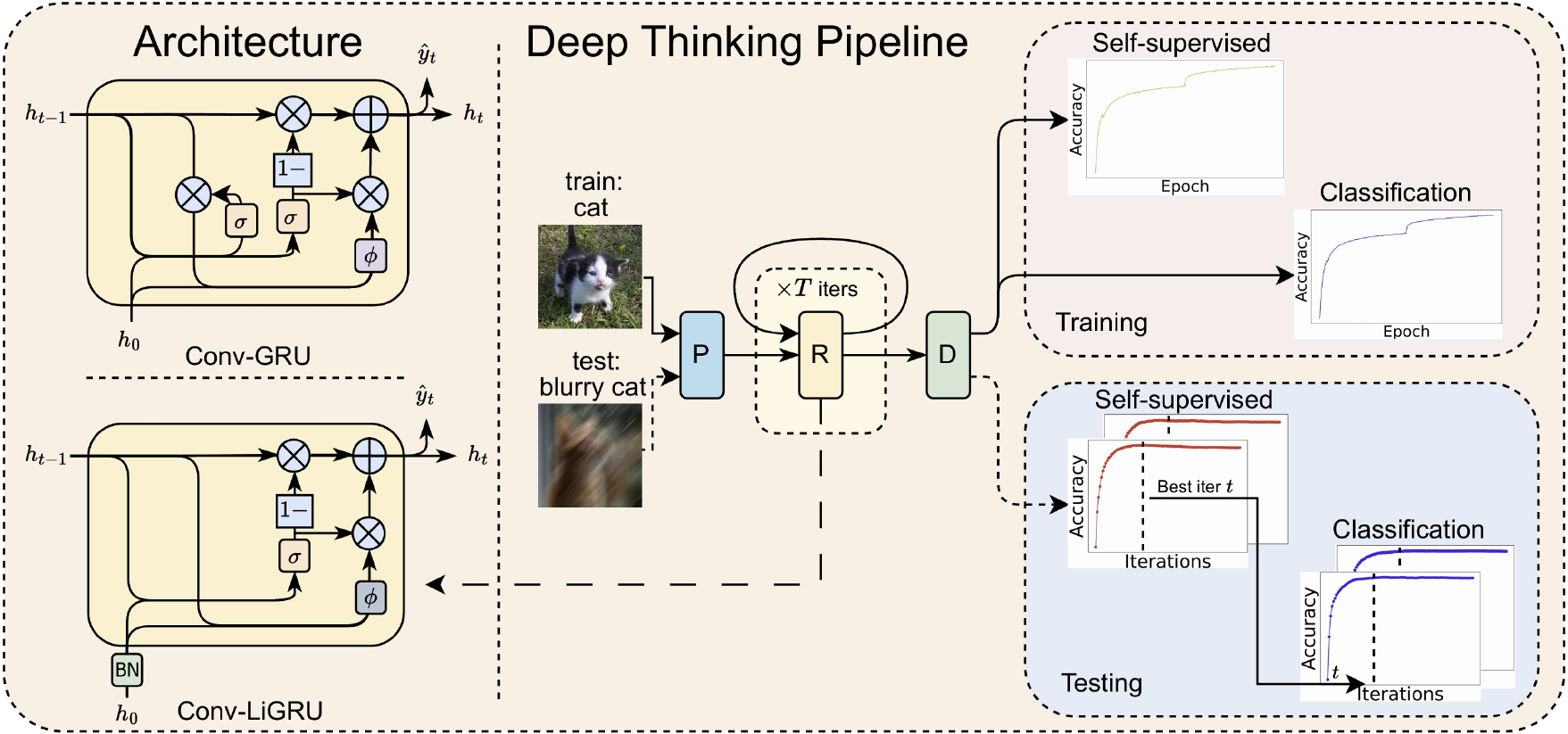}
    \caption{Deep thinking pipeline with training includes both main (classification) and secondary (self-supervised) tasks. 
    During inference, the iteration with peak self-supervised performance is selected for the halting of the classification task. 
    The reset gate in Conv-GRU is simplified and $\phi$ is replaced from Tanh to ReLu, resulting in Conv-LiGRU.}
    \label{fig:enter-label}
\end{figure}

%% file: section/methodology.tex
\section{Methodology}
\subsection{Deep Thinking Model Overview}
We study the deep thinking network that processes input images $\mathbf{X} \in \mathbb{R}^{C \times H \times W}$ in three main stages to explore the behavior of recurrent models under adaptive computation and extrapolation:

\textbf{Input Transformation}: The input image $\mathbf{X}$ is transformed via a convolutional layer, producing the initial state $\mathbf{h}_0$:
   \begin{equation}
       \mathbf{h}_0 = \sigma(\mathbf{W}_\text{in} \ast \mathbf{X} + \mathbf{b}_\text{in}),
   \end{equation}
where \( \mathbf{W}_\text{in} \) and \( \mathbf{b}_\text{in} \) are the weights and bias of the convolution, \( \ast \) denotes the convolution operator, and \( \sigma(\cdot) \) is an activation function such as ReLU.
\label{input_transform}

\textbf{Thinking Processing}: The thinking process employs a recurrent layer with identical input and output shapes, allowing the model to iteratively apply this layer multiple times, similar to an RNN.
With this architecture, the model can adjust the number of iterations based on the complexity of the task, a process we refer to as "thinking". We formulate this architecture as follows:
\begin{equation}
   \mathbf{h}_t = f_\text{rec}(\mathbf{h}_{t-1}), \quad t = 1, \dots, T_\text{train}
\end{equation}
where \( f_\text{rec}(\cdot) \) represents the recurrent function, and $T_{train}$ is
the maximum number of "thinking" steps during training. 

To ensure that the model does not "forget" the initial input, \citet{bansal2022endtoend} proposed Recall, which concatenates the input \( X \) with \( \textbf{h}_{t-1} \) at each "thinking" step, resulting in \( \textbf{h}_t = f_{\text{rec}}([\textbf{h}_{t-1}, X]) \). 
However, this approach requires the hidden feature map size to match the input \( X \), increasing computational complexity. 
Instead, we integrate \( \textbf{h}_{t-1} \) and \( \textbf{h}_0 \) at each iteration, with \( f_{\text{rec}} \) designed as a recurrent unit, leading to \( \textbf{h}_t = f_{\text{rec}}(\textbf{h}_{t-1}, \textbf{h}_0) \). 
The hidden feature map \( \textbf{h}_i \) can be downsampled compared to \( X \), reducing computational complexity compared to Recall. 
In this study, we experiment with different architectures for \( f_{\text{rec}} \) and propose a novel design introduced in Section~\ref{subsec:arch}.




\textbf{Output Prediction}: The output at each iteration \( \hat{y}_t \) is generated by applying a readout function:
   \begin{equation}
       \hat{\mathbf{y}}_t = f_\text{out}(\mathbf{h}_t),
   \end{equation}
   where \( f_\text{out}(\cdot) \) maps the recurrent state to the desired task output.
\label{output_pred}

By constraining a consistent target at each iteration, these stages enable us to examine the effect of varying the number of iterations $t$ on the model's performance, highlighting the adaptability and extrapolation capacity of recurrent models.

\subsection{Accuracy-Iteration Relationship Estimation}

Since ground-truth labels are unavailable during testing, assessing the performance on the main task ($\mathcal{T}_\text{main}$) across iterations ($t$) is nontrivial. 
To address this, we introduce a simple self-supervised auxiliary task ($\mathcal{T}_\text{aux}$) as a proxy, leveraging its strong correlation with $\mathcal{T}_\text{main}$ to estimate main task's performance.
The auxiliary task should satisfy the following assumptions: 

\textbf{Core Assumptions}:  
1. \( \mathcal{T}_\text{aux} \) shares semantic and structural similarities with \( \mathcal{T}_\text{main} \), ensuring a positive correlation between their accuracies over \( t \):  
   \[
   \textnormal{corr}\left(\text{Accuracy}_{\mathcal{T}_\text{aux}}(t), \text{Accuracy}_{\mathcal{T}_\text{main}}(t)\right) > 0 \ \forall\ t
   \]
2. The difficulty of \( \mathcal{T}_\text{aux} \) positively correlates with \( \mathcal{T}_\text{main} \) under both in-distribution (ID) and out-of-distribution (OOD) conditions. 

\textbf{Auxiliary Task Design}: 
We use the rotation prediction task \cite{Balaji2018MetaRegTD} as \( \mathcal{T}_\text{aux} \), a simple and effective pretext task that allows the model to learn features that are useful for many downstream tasks. 
Furthermore, we hypothesize that similar to the recognition task, rotation prediction becomes more difficult under more challenging conditions. 
The input image \( X \) is rotated by one of four angles \( \{0^\circ, 90^\circ, 180^\circ, 270^\circ\} \), and the model predicts the rotation angle as a four-class classification problem.
The auxiliary loss with $t$ iteration is:
\[
\mathcal{L}_t^{\text{aux}} = -\sum_{k=1}^{4} \mathbf{y}_t^{\text{a}}[\text{k}] \log \hat{\mathbf{y}}_t^{\text{a}}[\text{k}],
\]
where $\mathbf{y}_t^{\text{a}}[\text{k}]$ is the probability for each angle.

During training, at each update, we use the output of the last iteration $T_{train}$ to calculate the loss function. 
Combining with the main task loss \( \mathcal{L}_{T_{train}}^\text{main} \), which is the cross-entropy loss in classification, we obtain the total training loss:
\begin{align}
    \mathcal{L} = \mathcal{L}_{T_{train}}^\text{main} + \mathcal{L}_{T_{train}}^\text{aux}.
\end{align}

\textbf{Iteration Search for Main Task Improvement}:  
During testing, \( \text{Accuracy}_{\mathcal{T}_\text{aux}}(t) \) is used to estimate the optimal iteration \( t_\text{opt} \) for \( \mathcal{T}_\text{main} \). 
Given a fixed budget $T_{\text{test}}$,
\begin{align}
    t_\text{opt} = \underset{{t\in[T_{\text{test}}]}}{\arg\max} \big(\text{Accuracy}_{\mathcal{T}_\text{aux}}(t)\big).
\end{align}
Hence, comparing to the theoretical optimal accuracy given $t_*$, i.e., $A_*=\text{Accuracy}_{\mathcal{T}_\text{train}}(t_*)$, our hypothesis estimates the optimal accuracy as
\begin{align}
    \hat{A}_* = \text{Accuracy}_{\mathcal{T}_\text{main}}(t_\text{opt}).
\end{align}

This framework offers a scalable and adaptive method for estimating the accuracy of the main task and optimizing performance efficiently in diverse test scenarios.
The inference pipeline is described at Algorithm \ref{alg:estimate_acc}.

\subsection{Conv-LiGRU}
\label{subsec:arch}
\begin{algorithm}[t!]
\caption{Testing Phase: Accuracy-Iteration Relationship Estimation}
\begin{algorithmic}[1]
\State \textbf{Input:} Trained model, test data with $D$ batches, maximum iterations $T_\text{test}$
\State \textbf{Output:} Optimal iteration $t_\text{opt}$ 

\Statex
\State \textbf{Step 1: Initialize variables}
\State Initialize \texttt{Correct} as a zero array of length $T_\text{test}$
\Statex

\State \textbf{Step 2: Compute $\text{Accuracy}_{\mathcal{T}_\text{aux}}(t)$ during testing}
\For{$i = 1, 2, \dots, D$} \Comment{Iterate over $D$ test batches}
    \State $h^{(0)} \gets \text{InputTransformation}(\text{Batch}_i)$
    \For{$t = 1, 2, \dots, T_\text{test}$} \Comment{Iterate over $T_\text{test}$ steps}
        \State $h_t \gets f_\text{rec}(h_{t-1}, h_0)$ \Comment{Recurrent computation}
        \State $\hat{y}_t^{\text{main}}, \hat{y}_t^{\text{aux}} \gets f_\text{out}(h_t)$ \Comment{Output predictions}
        \State $\texttt{Correct}[t] \mathrel{+}=  \text{\# Correct auxiliary samples}$
    \EndFor
\EndFor
\State $\text{Accuracy}_{\mathcal{T}_\text{aux}}(t) \gets \texttt{Correct}[t] / \text{TotalSamples}$
\Statex

\State \textbf{Step 3: Estimate optimal iteration}
\State $t_\text{opt} \gets \arg\max_t \big(\text{Accuracy}_{\mathcal{T}_\text{aux}}(t)\big)$

\Statex
\State \textbf{Return:} $t_\text{opt}$
\end{algorithmic}
\label{alg:estimate_acc}
\end{algorithm}

Existing research has explored various recurrent architectures for image input, including Recursive Convolutional Networks \cite{recursiveCnn2014lecun}, GRU-based models \cite{Ballas2016Iclr}, and Deep Thinking Recall. 
However, these architectures often fail to achieve stable accuracy across iterations or miss the long term memory.

In this study, we propose a novel GRU-based model, Conv-LiGRU, inspired by the light-gated recurrent unit (LiGRU) \cite{ligru2018}, which simplifies the GRU by removing the reset gate, replacing tanh with ReLU activation, and applying batch normalization. 

\begin{figure*}[t!]
    \centering
    \includegraphics[width=0.95\linewidth]{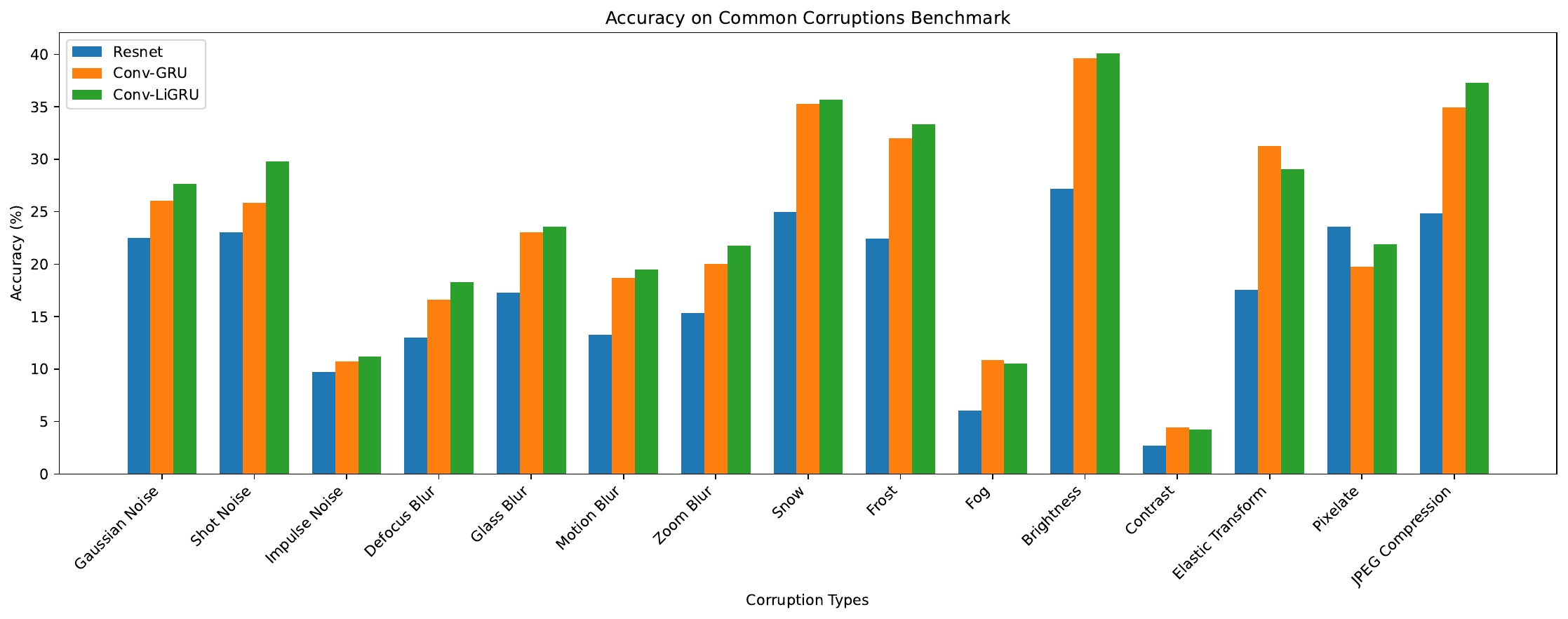}
        \caption{Accuracy (\%) on CIFAR100-C, at level 5, Resnet, Conv-GRU, Conv-LiGRU}
        \label{fig:bar_chart_cifar100c}
\end{figure*}

Key features of Conv-LiGRU include:

\textbf{Removal of the Reset Gate}: By eliminating the reset gate, Conv-LiGRU streamlines the gating mechanism, reducing the number of parameters, and improving computational efficiency. LiGRU was originally designed for audio data, where the authors argued that the reset gate in GRU might disrupt intermediate features, particularly for continuous data like audio. 
Similarly, the reasoning process consists of a sequence of thinking steps, where skipping even a single step can lead to incorrect conclusions. Therefore, removing the reset gate is a reasonable choice to ensure a stable flow of information and a consistent, uninterrupted thinking process.


\textbf{Normalization and Activation Function}: Similar to LiGRU, Conv-LiGRU replaces \( \tanh \) with ReLU to mitigate the vanishing gradient problem and enhance the model's ability to capture long-range dependencies. Additionally, instead of layer normalization, Conv-LiGRU employs batch normalization \cite{he2016residual} to ensure stable performance during iterative computations and to better suit image-based data and convolution operations.

\textbf{Convolutional State Transitions}: 
Conv-LiGRU adapts LiGRU for image tasks by replacing fully connected layers with convolutions, preserving spatial structure and enhancing performance.

The state update equation for Conv-LiGRU is defined as:
   \begin{equation}
       \textbf{h}_t = \textbf{z}_t \odot \tilde{\textbf{h}}_t + (1 - \textbf{z}_t) \odot \textbf{h}_{t-1},
   \end{equation}
   \begin{equation}
       \textbf{z}_t = \sigma(U_z \ast \textbf{h}_{t-1} + BN(W_z \ast \textbf{h}_0)),
   \end{equation}
   \begin{equation}
       \tilde{\textbf{h}}_t = RELU(BN(W_h \ast \textbf{h}_0) + U_h \ast \textbf{h}_{t-1})
   \end{equation}
where \( z_t \) is the update gate, \( \tilde{h}_t \) is the candidate state, and \( \odot \) denotes element-wise multiplication, \( \ast \) denotes the convolution operation, and \( W_z \), \( W_h \), \( U_z \), and \( U_h \) are learnable parameters, $RELU$, $BN$ are relu activation function and bach normalize layer perspective.

These enhancements allow Conv-LiGRU to maintain stability and achieve robust performance across ID and OOD scenarios. 
Reduced computational complexity and enhanced stability make Conv-LiGRU highly effective for iterative image-based tasks.

%% file: section/experiment.tex
\section{Experiment}
\subsection{Datasets}  
We train our models on the CIFAR10 and CIFAR100 and evaluate them using CIFAR10-C and CIFAR100-C datasets.  

\textbf{CIFAR10 \& CIFAR100} 
are standard benchmarks for image classification. 
CIFAR10 has 50,000 training and 10,000 testing images across 10 classes, while CIFAR100 spans 100 classes with the same dataset size. 
We use both for training and validation.  
\\
\textbf{CIFAR10-C \& CIFAR100-C} 
are corrupted versions of CIFAR10 and CIFAR100. 
They are used to assess models' robustness to distribution shifts, featuring 15 corruption types (e.g., Gaussian noise, blur, contrast) at five severity levels. 
We use them to evaluate generalization under real-world distortions.
\subsection{Models and Training}
\textbf{Model Architecture.} 
\begin{figure}[t!]
    \centering
    \includegraphics[width=0.85\linewidth]{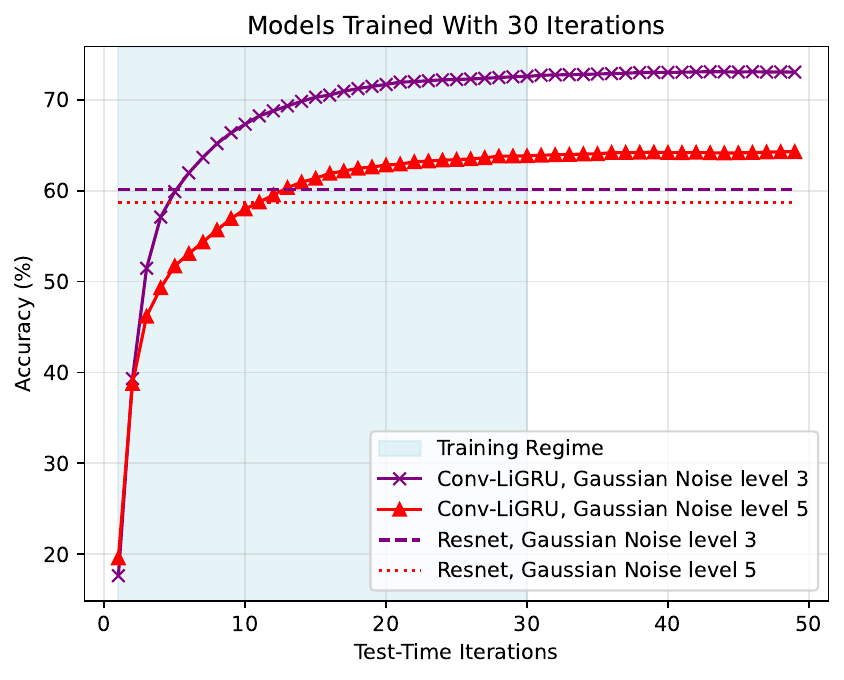}
    \caption{We evaluate the ability of networks to classify objects on two test sets with higher noise severity than those used during training, measuring accuracy on level 3 inputs (purple) and level 5 inputs (red). Recurrent models are compared against the best feed-forward models.}
    \label{fig:visualize-adaptive}
\end{figure}
We kept the Image Transformation and Output Prediction components (Section~\ref{output_pred}) to evaluate extrapolation, varying the Thinking Processing between feed-forward and recurrent architectures.

We used ResNet \citep{he2016residual} with 4 layers, each layer includes 6 convolution blocks, a total of 24 convolutional blocks for the feed-forward model, maintaining equal input-output channels. 
We compare 3 main architectures: Recall architecture, Conv-GRU, and Conv-LiGRU. 
All models used 128 feature channels, and recurrent models were trained for \( T_{\text{train}} = 30 \) iterations, for testing we use the $T_{test} = 100$.  

Training used the Adam optimizer \citep{adam} with a weight decay of 0.0002. 
Datasets were split 80\%/20\% for training and validation. 
Following \citet{aug_cifarc}, Gaussian noise\todo[disable]{what does $\sigma$ means? Is it the variance or the weight of the noise?} \todo[disable]{Fixed} was added to enhance generalization. For self-supervised learning,  each input image was randomly rotated by the function $f_{rotate}$ at one of four angles (0°, 90°, 180°, 270°), so the final augmentation as: 
\begin{equation}
    x = f_{rotate}(clip(x + \delta))
\end{equation}
where $\delta \sim \mathcal{N}(0, \sigma^2 \mathbf{I})$, $\sigma$ is the standard deviation of the Gaussian noise, and $x + \delta$ is clipped to the input range $[0, 1]^N$. We set $\sigma = 0.04$ equivalent to level-1 Gaussian noise corruption in CIFAR10-C.

We trained these models for 200 epochs on CIFAR10 and for 600 epochs on CIFAR100.  

\subsection{Extrapolation Capability of Recurrent Models}
\begin{figure}[t!]   
    \centering
    \includegraphics[width=\linewidth]{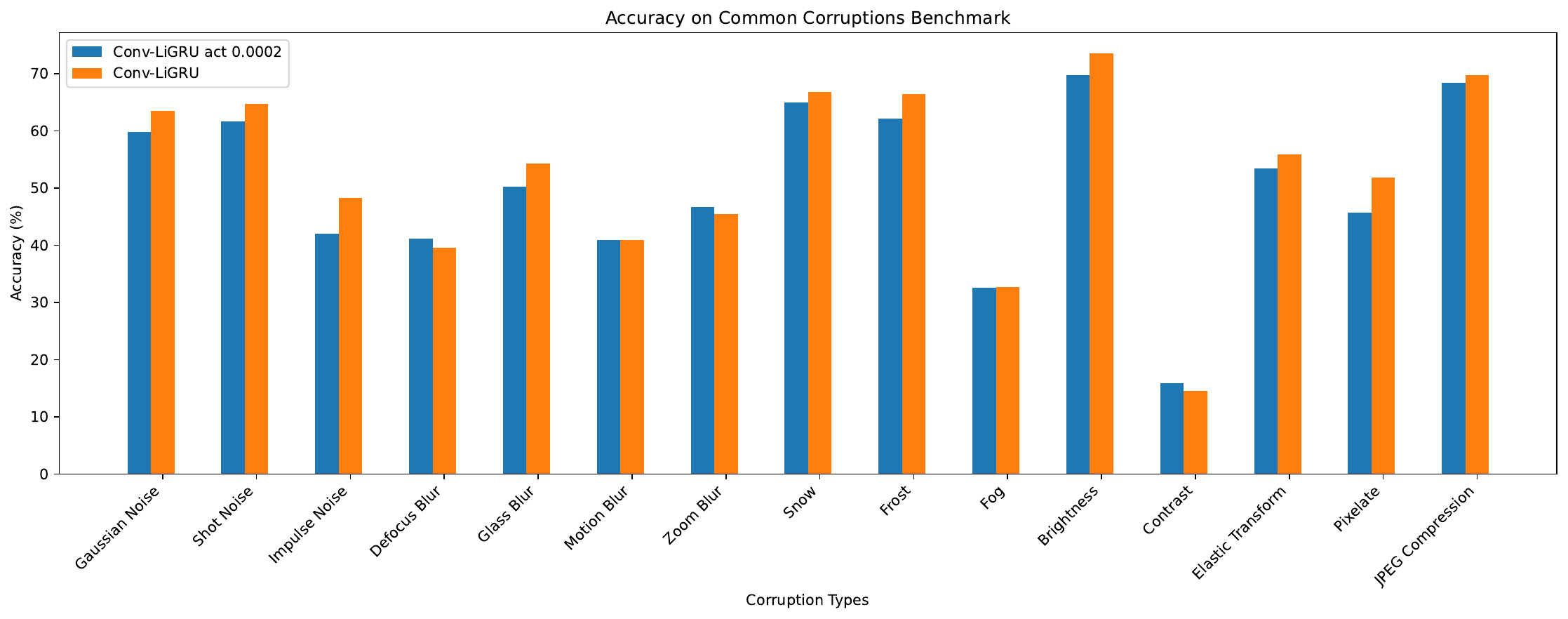}
        \caption{Accuracy (\%) on CIFAR10-C, at level 5, Conv-LiGRU with and without ACT}
        \label{fig:bar_chart_act_ligru}
\end{figure}
Recurrent networks are well-known for their generalization capabilities in logical tasks. 
In our study, we extend this observation to computer vision tasks. 
By training various recurrent models on clean datasets augmented from CIFAR10 and CIFAR100, and testing them on 15 corruption types from CIFAR10-C and CIFAR100-C, we found that recurrent models consistently outperform feed-forward networks in generalization. 

Unlike feed-forward networks, which operate with fixed computation costs, recurrent models adapt dynamically, iterating more when encountering challenging samples with higher corruption levels. 
This "thinking deeper" ability enhances their generalization performance. 
Figure \ref{fig:bar_chart_cifar100c} highlights this, showing that Conv-GRU and Conv-LiGRU outperform ResNet on most corruption test sets.
More results with CIFAR-10C are demonstrated in Appendix~\ref{sec:addition_result}.

Recurrent models also demonstrate remarkable parameter efficiency, as shown in Table \ref{tab:params}, using only one-sixth the parameters of feed-forward networks while achieving superior generalization. 
Figure \ref{fig:visualize-adaptive} further illustrates their adaptability. 
For example, at noise level 3, the model requires just 6 iterations to achieve 60\% accuracy, while at noise level 5, it needs 13 iterations to achieve the same accuracy. 
Notably, a recurrent model with 24 iterations outperforms a feed-forward network with equivalent computational depth, emphasizing the effectiveness of dynamic depth adjustment. 

Beyond adapting computation during inference, recurrent models learn invariant features that are reusable across iterations. 
This synergy of dynamic computation and feature reuse makes them powerful for tackling complex computer vision tasks.

\subsubsection{Iterative Outputs}
\begin{figure}[h]
    \centering
    \includegraphics[width=1\linewidth]{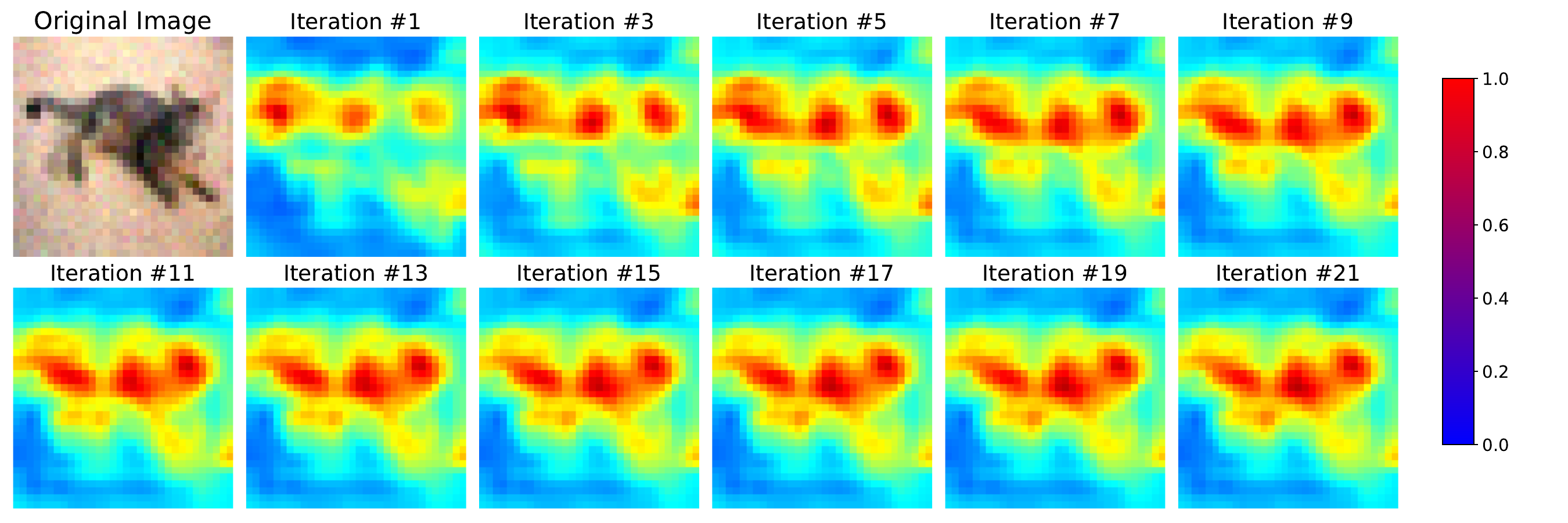}
    \caption{A "Cat" sample input, and outputs from different iterations are shown to illustrate the model’s sequential reasoning process on CIFAR10-C (level 1). We visualize the norm of vector feature \( h^{(i, j)}_t \) (row \( i \), column \( j \)) of the feature map \( h_t \), demonstrating the model’s feature extraction over iterations. This is a representative example from a Conv-LiGRU model trained on CIFAR10 with \( T_{\text{train}} = 30 \).}
    \label{fig:hidden_states}
\end{figure}

To understand the model's thinking process, we visualize the hidden feature maps \( h_t \) at each iteration. Figure \ref{fig:hidden_states} presents the Gaussian heatmap of the norm vector feature \( h^{(i, j)}_t \) (at row \( i \), column \( j \)) for each feature map \( h_t \). This figure reveals two notable insights. First, the model progressively detects features from local to global, gradually capturing the entire object. Second, it prioritizes identifying key distinguishing features, such as the face and tail, before detecting less critical ones, like the legs. This suggests that the model’s thinking process can be reasoned about, and it exhibits a naive yet intuitive recognition process similar to human perception.

\begin{table}[t!]
\centering
\caption{The number of parameters}
\begin{tabular}{l|cc}
\hline
\textbf{Model} & \textbf{Params} & \textbf{Compress} \\ \hline
Resnet & 7.8M & $1.0\times$\\
Recall & 0.9M & $8.3\times$ \\
Conv-GRU & 1.6M & $4.9\times$ \\
Conv-LiGRU & 1.3M & $6.0\times$ \\ \hline
\end{tabular}
\label{tab:params}
\end{table}

\subsection{Analysis Of The Underthinking Problem}
\todo[disable]{change the problem's name to underthinking}

ACT aims to optimize the iteration count in recurrent models but often limits their reasoning ability. 
Overemphasis on the "ponder cost" can cause RNNs to halt prematurely, effectively reducing them to feedforward networks. 
To examine the impact of adaptive computation, we incorporated the ACT mechanism into Conv-GRU and Conv-LiGRU and compared its performance with our proposed self-supervised accuracy estimation method.  

\begin{figure*}[htbp]
    \centering
    \begin{subfigure}{0.32\textwidth}
        \includegraphics[width=\linewidth]{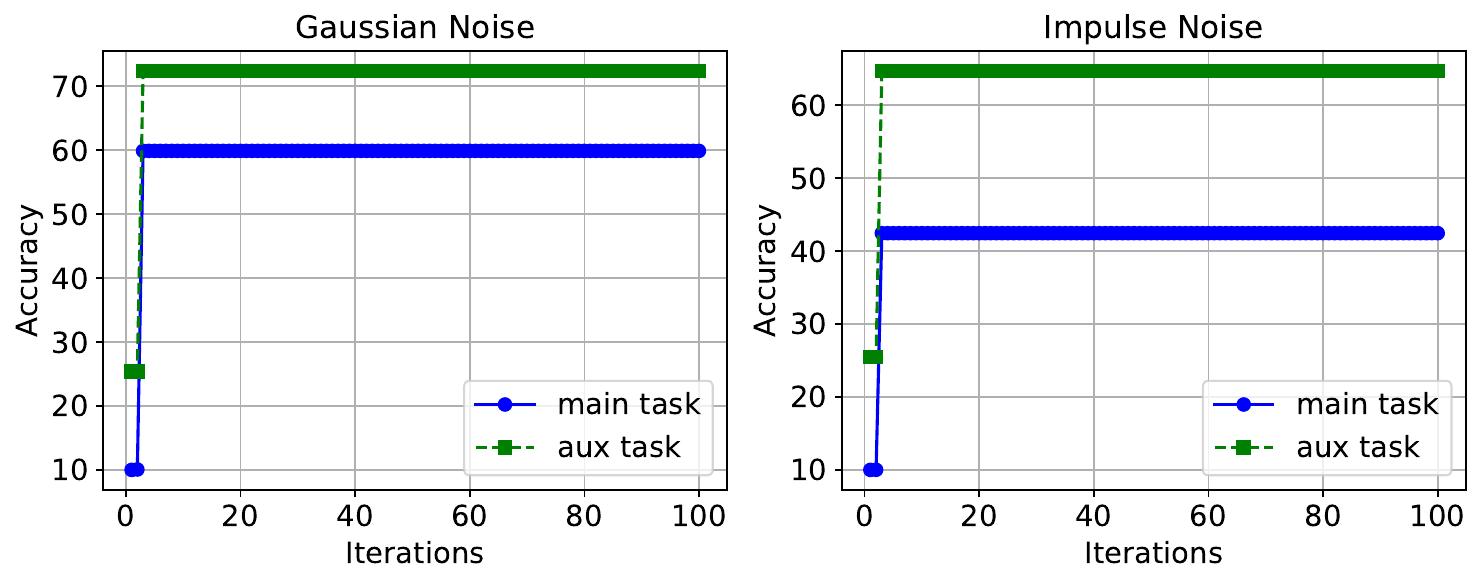}
        \caption{Conv-GRU with ACT, $\tau = 0.5$ }
        \label{fig:cnn_gru_gn_act_0.5}
    \end{subfigure}
    \hfill
    \begin{subfigure}{0.32\textwidth}
        \includegraphics[width=\linewidth]{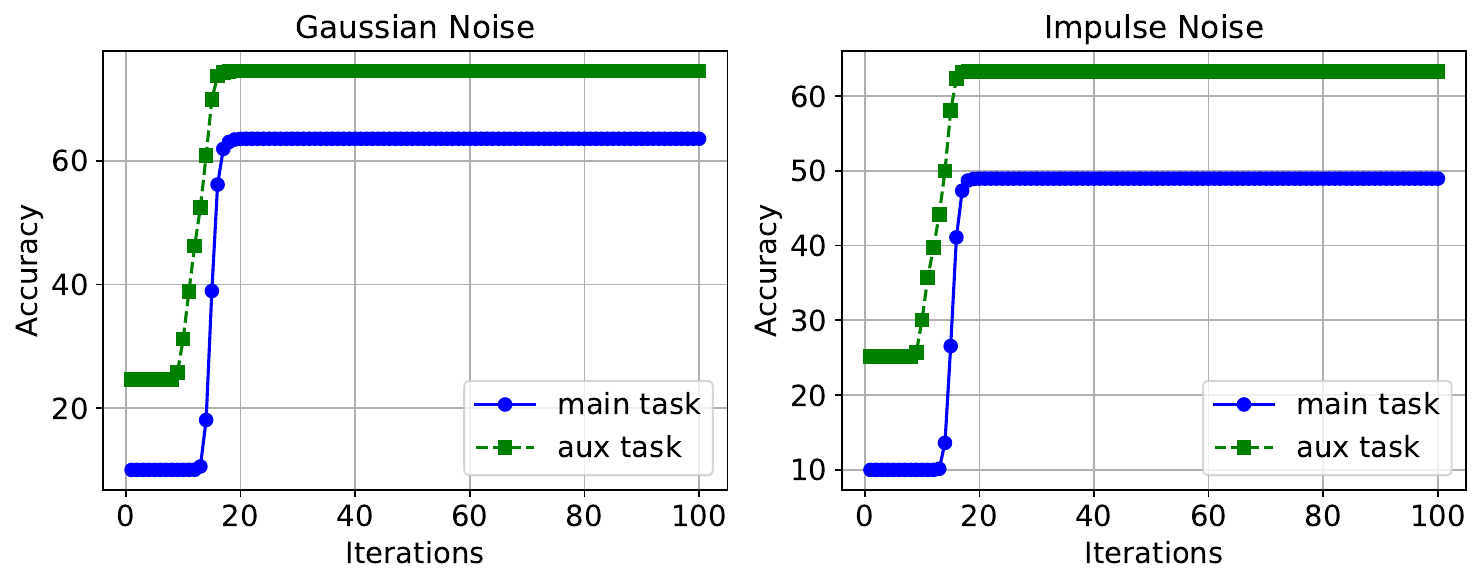}
        \caption{Conv-GRU with ACT, $\tau = 0.0002$ }
        \label{fig:cnn_gru_gn_act_0.0002}
    \end{subfigure}
    \hfill
    \begin{subfigure}{0.32\textwidth}
        \includegraphics[width=\linewidth]{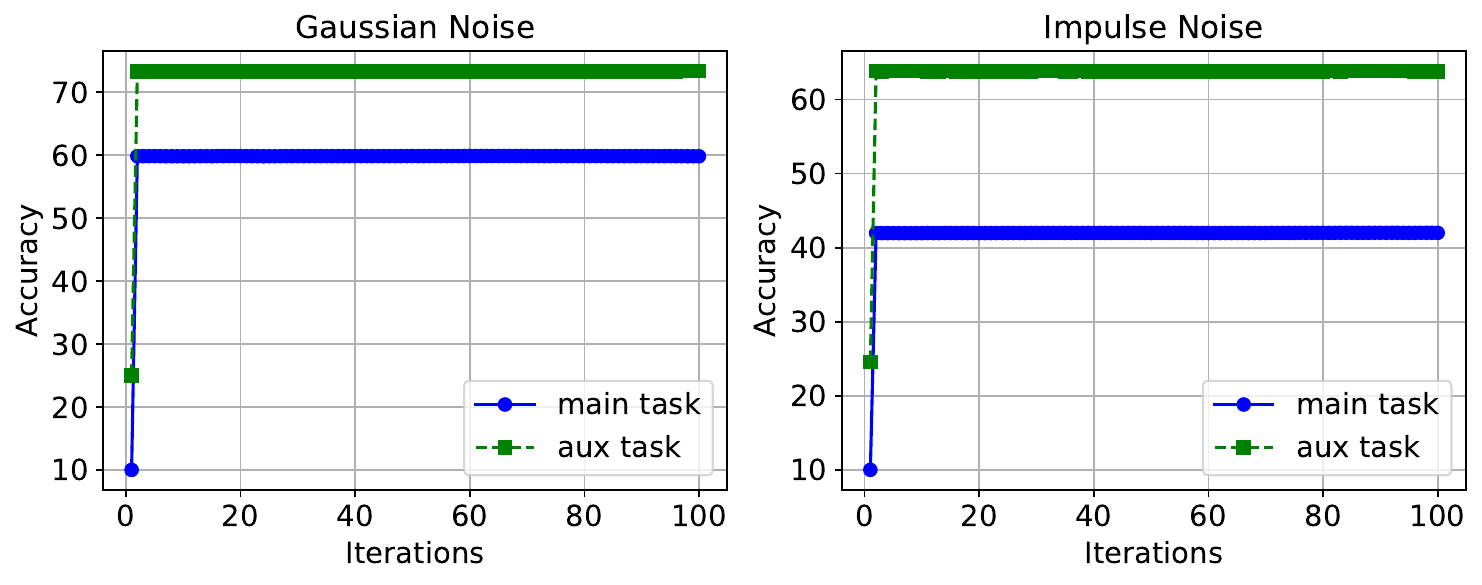}
        \caption{Conv-LiGRU with $\tau = 0.0002$}
        \label{fig:cnn_ligru_act_0.0002}
    \end{subfigure}
    \vfill
    \begin{subfigure}{0.32\textwidth}
        \includegraphics[width=\linewidth]{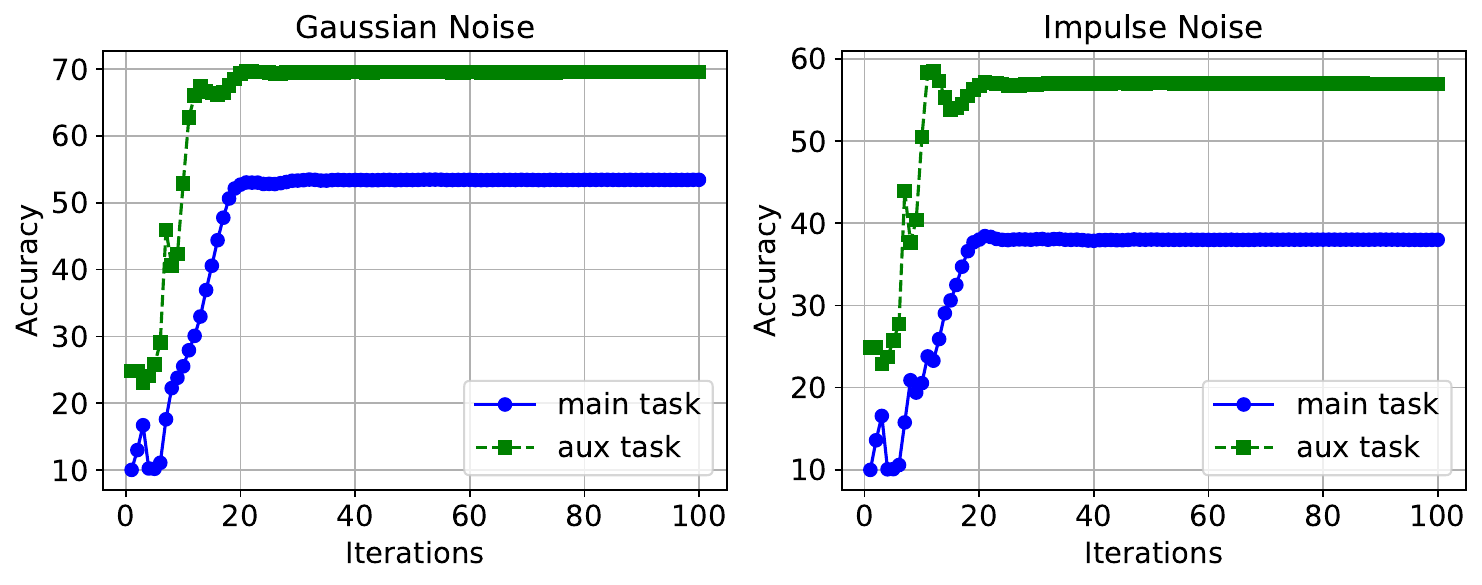}
        \caption{Recall}
        \label{fig:dt_net_recall_2d_alpha_0.0}
    \end{subfigure}
    \hfill
    \begin{subfigure}{0.32\textwidth}
        \includegraphics[width=\linewidth]{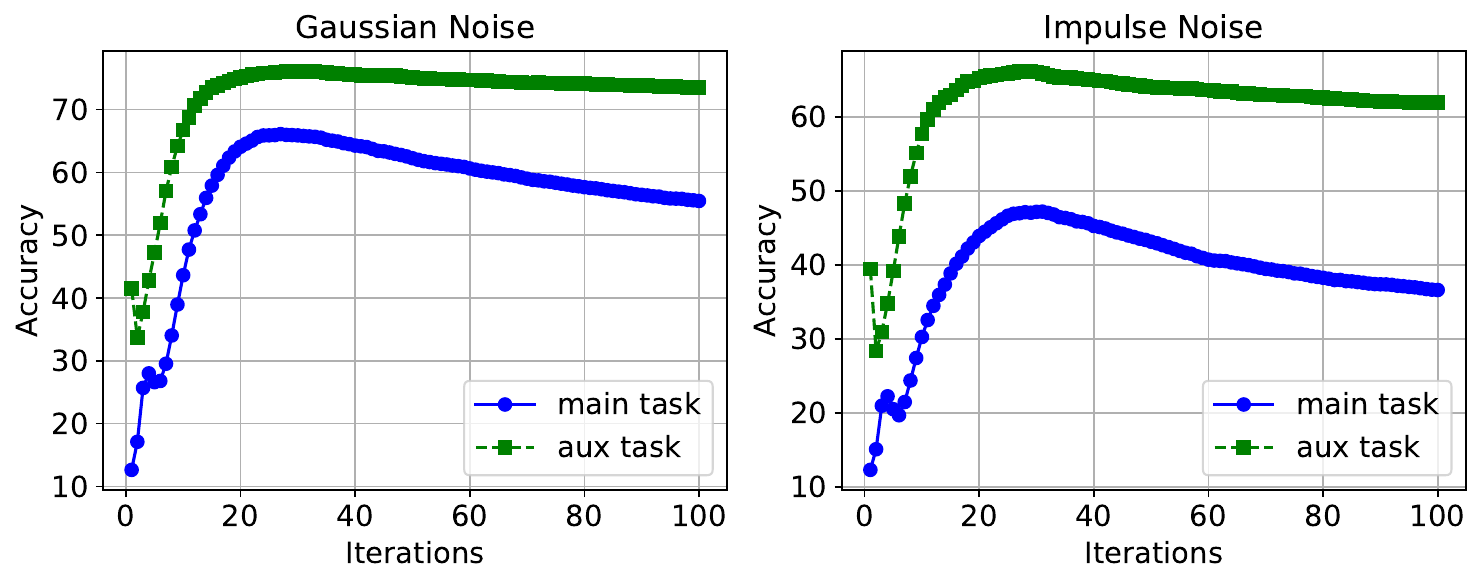}
        \caption{Conv-GRU}
        \label{fig:cnn_gru_gn_alpha_0.0}
    \end{subfigure}
    \hfill
    \begin{subfigure}{0.32\textwidth}
        \includegraphics[width=\linewidth]{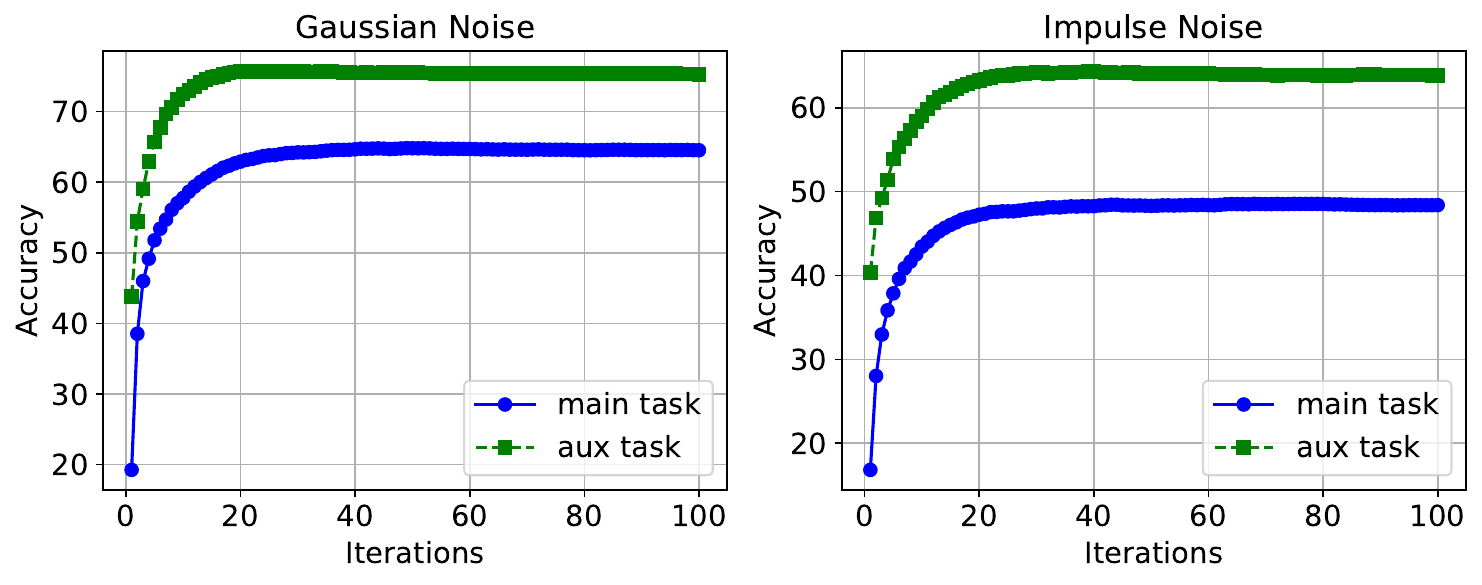}
        \caption{Conv-LiGRU}
        \label{fig:cnn_ligru_alpha_0.0}
    \end{subfigure}
    
    \caption{Accuracy across iterations of the diverse models on level 5 of corruption CIFAR10-C test sets.}
    \label{fig:arch_and_hyper}
\end{figure*}

We trained Conv-GRU model by the ACT method with different values for the hyperparameters $\tau$ and $\epsilon$, where $\tau$ is the weight of the "ponder cost" term in the loss function. A larger $\tau$ encourages the model to minimize the number of "thinking" steps during training. 
Additionally, if the cumulative stopping probability at each "thinking" step exceeds $1 - \epsilon$, the model terminates the thinking process. Therefore, a smaller $\epsilon$ leads to a longer thinking process. 
We provide a more detailed explanation of ACT and the hyperparameters $\tau$ and $\epsilon$ in the Appendix \ref{sec:background}. 
With $\tau = 0.5$ and $\epsilon = 1\text{e-5}$, the model halted after the third iteration (Figure \ref{fig:cnn_gru_gn_act_0.5}). 
Lowering $\tau$ to $2\text{e-4}$ extended its iterations to 18 (Figure \ref{fig:cnn_gru_gn_act_0.0002}). 
However, applying the same parameters to Conv-LiGRU still led to immediate halting after 2 iteration (Figure \ref{fig:cnn_ligru_act_0.0002}).

This early termination significantly restricts the extrapolation capabilities of recurrent models. 
Instead, we propose allowing models to compute adaptively with an upper limit defined by $T_{test}$ and estimating the optimal iteration ($t_{opt}$) via a self-supervised task. 

\begin{table}[t!]
\setlength{\tabcolsep}{3.5pt} 
\centering
\caption{The average estimated accuracy and maximum accuracy across all 15 types of corruption at level 5 in the CIFAR10-C dataset. ($\tau_1 = 0.5, \tau_2=0.0002$ are the values of $\tau$ when we apply ACT)}
\begin{tabular}{l|cc|cc}
\hline
\textbf{Model} & \textbf{Est \%} & \textbf{Max \%} & \textbf{Avg $t_{opt}$} & \textbf{Var $t_{opt}$} \\ \hline
resnet & 39.58 & 39.58 & 24.0 & 0.0 \\
recall & 42.76 & 44.43 & 21.0 & 90.0 \\
conv-gru, $\tau_1$ & 49.89 & 49.89 & 3 & 0.0 \\
conv-gru, $\tau_2$ & 50.60 & 50.60 & 17.8 & 0.45 \\
conv-ligru, $\tau_2$ & 50.42 & 50.42 & 2.0 & 0.06 \\
conv-gru & 51.47  & 52.50 & 19.6 & 12.97 \\
conv-ligru & \textbf{52.54} & \textbf{52.89} & 23.6 & 38.11 \\ \hline
\end{tabular}
\label{tab:acc_gap}
\end{table}

Our approach allows recurrent models to "think" freely, leading to notable performance improvements. 
From Table~\ref{tab:acc_gap}, we observe that Conv-GRU and Conv-LiGRU, when not constrained by ACT, tend to take more thinking steps and achieve superior performance compared to when ACT restricts them. Figure~\ref{fig:bar_chart_act_ligru} also demonstrates that the estimation of optimal iterations consistently outperforms the computation constraint by ACT, achieving superior results on almost 15 corruption types in CIFAR10-C.

\subsection{Efficiency of Conv-LiGRU}
\textbf{Conv-LiGRU Mitigates Overthinking.} 
Figure~\ref{fig:cnn_gru_gn_alpha_0.0} shows that Conv-GRU suffers from overthinking on level-5 corruption test sets in CIFAR10-C, affecting both main and auxiliary tasks.

To address this, Conv-LiGRU removes the reset gate to better retain information across iterations. 
This is crucial for handling corrupted data, where feature extraction is more challenging. 
Figure~\ref{fig:cnn_ligru_alpha_0.0} confirms that Conv-LiGRU significantly reduces overthinking across all corruption test sets, offering a robust solution.

While Recall demonstrates resistance to overthinking (Figure \ref{fig:dt_net_recall_2d_alpha_0.0}), it requires maintaining full feature map resolution, leading to higher computational costs. Furthermore, its lack of long-term memory results in inferior performance in CIFAR10-C compared to GRU-based models (Table \ref{tab:acc_gap}).

\textbf{Outstanding Performance of Conv-LiGRU.} Tables \ref{tab:acc_gap} demonstrate that Conv-LiGRU achieves higher estimated and peak accuracy across diverse models on the CIFAR10-C dataset. 

Figure \ref{fig:bar_chart_cifar100c} further illustrates Conv-LiGRU's superiority, outperforming Conv-GRU on 11 out of 15 corruption types at level 5 in CIFAR100-C. With an average accuracy improvement of $0.98\%$ (as shown in Table \ref{tab:acc_avg_cifar100}), these findings underscore Conv-LiGRU's remarkable effectiveness in handling image data and its clear advantage over Conv-GRU.

\subsection{Limitation of rotation prediction task}
\begin{figure}[!t]
    \centering
    \begin{subfigure}{0.1\textwidth}
        \includegraphics[width=\linewidth]{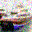}
        \caption{"Ship"}
        \label{fig:ship_sample}
    \end{subfigure}
    \vfill
    \begin{subfigure}{0.22\textwidth}
        \includegraphics[width=\linewidth]{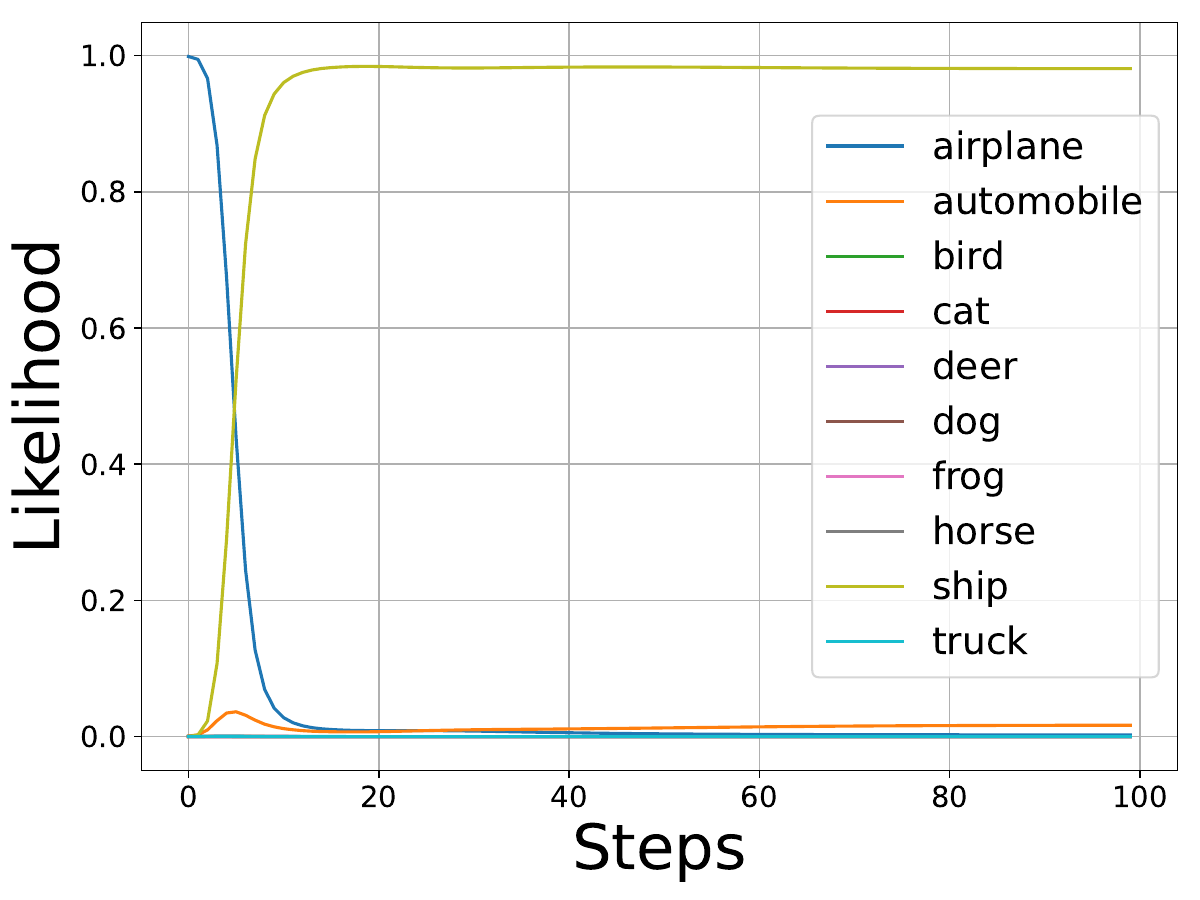}
        \caption{Likelihood of all 10 classes over steps}
        \label{fig:all_classes_main_likelihood_hard_ship}
    \end{subfigure}
    \begin{subfigure}{0.22\textwidth}
        \includegraphics[width=\linewidth]{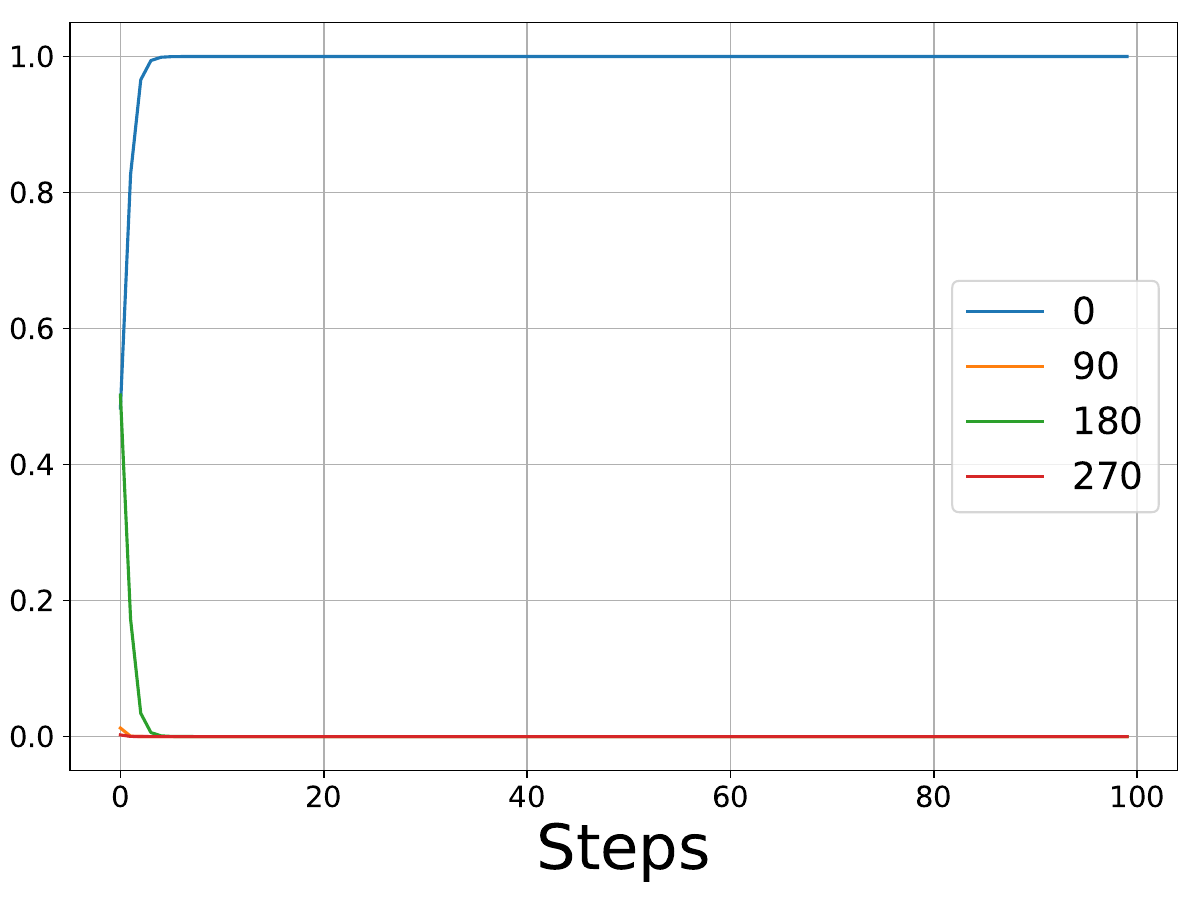}
        \caption{Likelihood of all 4 rotation classes over steps}
        \label{fig:all_class_likelihood_aux_hard_ship}
    \end{subfigure}
    \caption{Likelihood across iteration of a level 5 Gaussian Noise sample in class "Ship"}
    \label{fig:likelihood_ship}
\end{figure}
Figure \ref{fig:likelihood_ship} illustrates the likelihood of different classes across iterations, using a sample (Figure \ref{fig:ship_sample}) from the "Ship" class with a rotation angle of 0 degrees. The sample's features, which include many edge features, allow the model to predict rotation effectively and maintain stability over iterations (Figure \ref{fig:all_class_likelihood_aux_hard_ship}). This stability and accuracy in rotation prediction enable the recurrent model to better estimate the optimal iteration for the main task.

\begin{figure}[t!]
    \centering
    \begin{subfigure}{0.08\textwidth}
        \includegraphics[width=\linewidth]{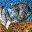}
        \caption{"Cat"}
        \label{fig:cat_sample}
    \end{subfigure}
    \vfill
    \begin{subfigure}{0.22\textwidth}
        \includegraphics[width=\linewidth]{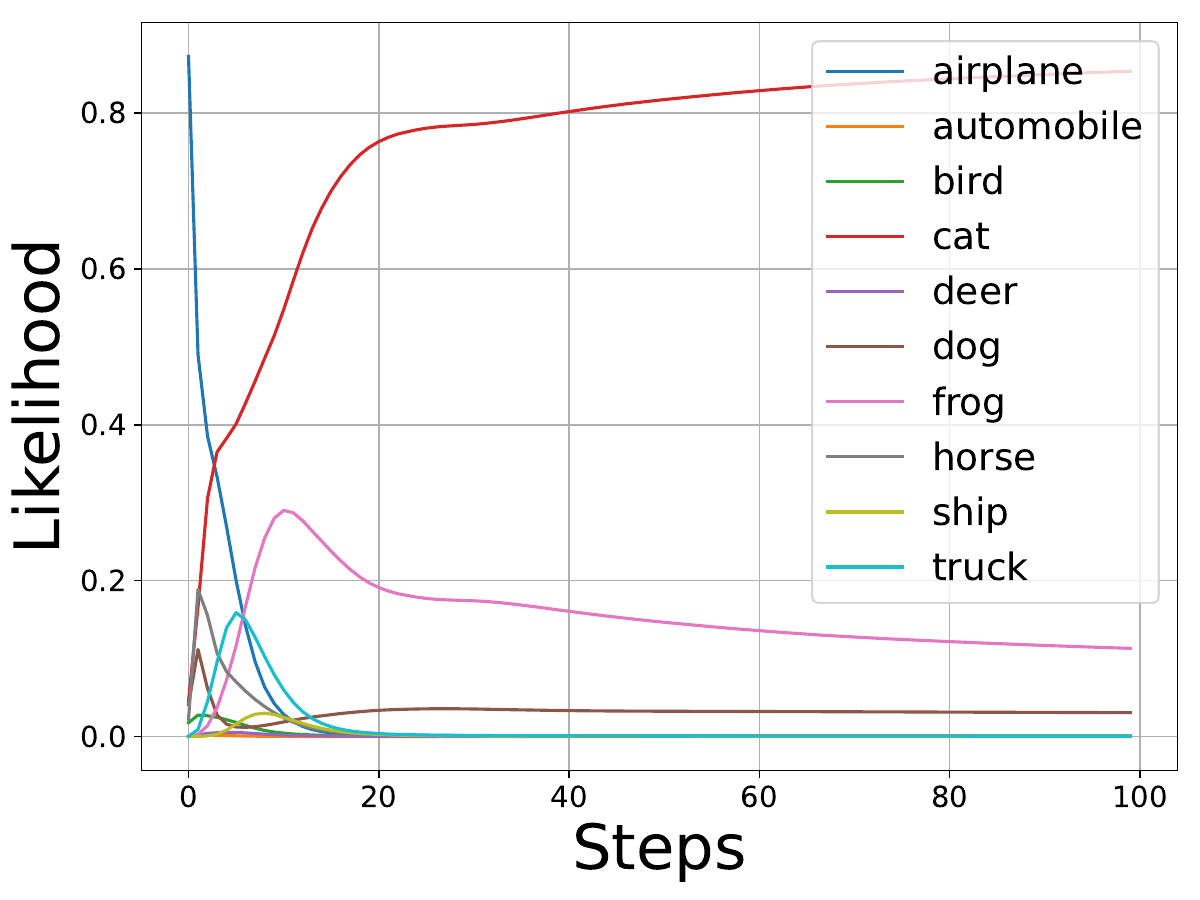}
        \caption{Likelihood of all 10 classes over steps}
        \label{fig:all_classes_main_likelihood_hard_cat}
    \end{subfigure}
    \begin{subfigure}{0.22\textwidth}
        \includegraphics[width=\linewidth]{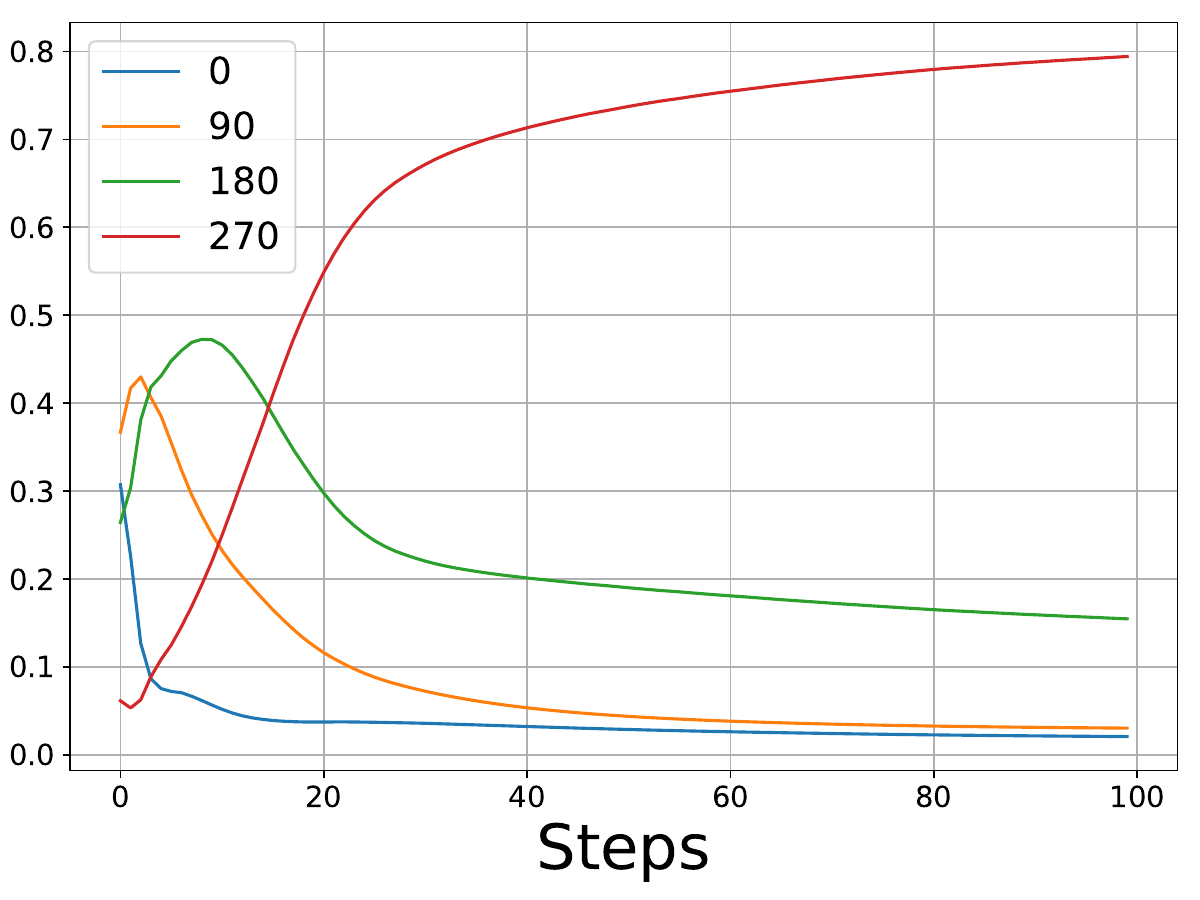}
        \caption{Likelihood of all 4 rotation classes over steps}
        \label{fig:all_class_likelihood_aux_hard_cat}
    \end{subfigure}
    \caption{Likelihood across iteration of a level 5 Gaussian Noise sample in class "Cat"}
    \label{fig:likelihood_cat}
\end{figure}

However, predicting the rotation angle becomes more challenging for samples with fewer edge features or isotropic characteristics, such as the sample from the "Cat" class in Figure \ref{fig:cat_sample}. Figure \ref{fig:all_classes_main_likelihood_hard_cat}, \ref{fig:all_class_likelihood_aux_hard_cat} shows that while the likelihood of the "Cat" class continues to increase over iterations and remains significantly higher than other classes, the likelihood of the ground truth rotation (0 degrees) in the self-supervised task is very low. Instead, the model tends to predict the class corresponding to a 270-degree rotation. This behavior negatively impacts the estimation of the optimal iteration for the main task based on the self-supervised task.

We acknowledge this as a limitation of using the rotation prediction task as a self-supervised task.

\subsection{Converge to a fixed point does not ensure to mitigate "overthinking"}

\begin{table}[t!]
\centering
\caption{The average estimated accuracy, and maximum accuracy across all 15 types of corruption at level 5 in the CIFAR100-C dataset.}
\begin{tabular}{l|ccc}
\hline
\textbf{Model} & \textbf{Est \%} & \textbf{Max \%} & \textbf{Gap \%} \\ \hline
Resnet & 17.56 & 17.56 & 0 \\
Conv-GRU & 23.27  & 23.78 & -0.51 \\
Conv-LiGRU & \textbf{24.25} & \textbf{24.99} & -0.74 \\ \hline
\end{tabular}
\label{tab:acc_avg_cifar100}
\end{table}

\cite{bansal2022endtoend} highlights the relationship between changes in feature maps across iterations and the issue of "overthinking".
they demonstrate that if \(\|h_t-h_{t-1}\|_2\) converges to $0$ in deep thinking models, the feature map reaches a fixed point, making later predictions unchanged and mitigating "overthinking."

\begin{figure}[t!]
    \centering
    \begin{subfigure}{0.3\textwidth}
        \includegraphics[width=\linewidth]{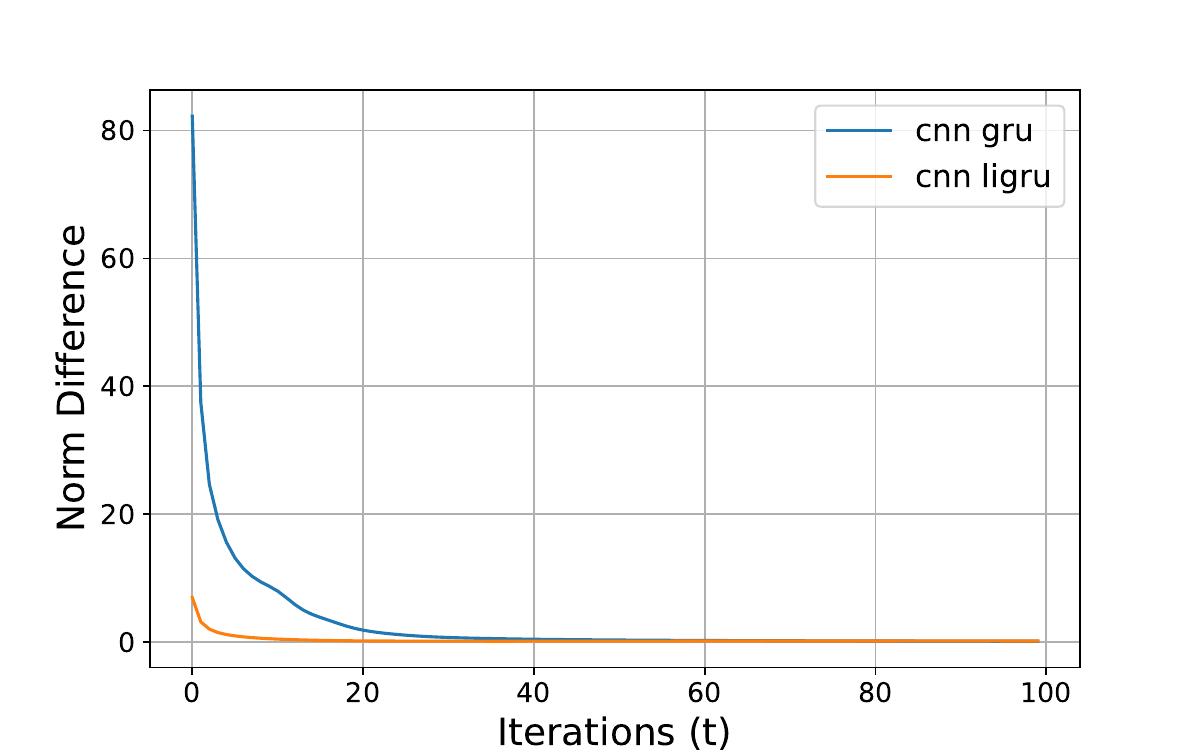}
        \caption{The $\|h_t - h_{t-1}\|_2$ across iterations}
        \label{fig:visualize norm}
    \end{subfigure}
    \vfill
    \begin{subfigure}{0.45\textwidth}
        \includegraphics[width=\linewidth]{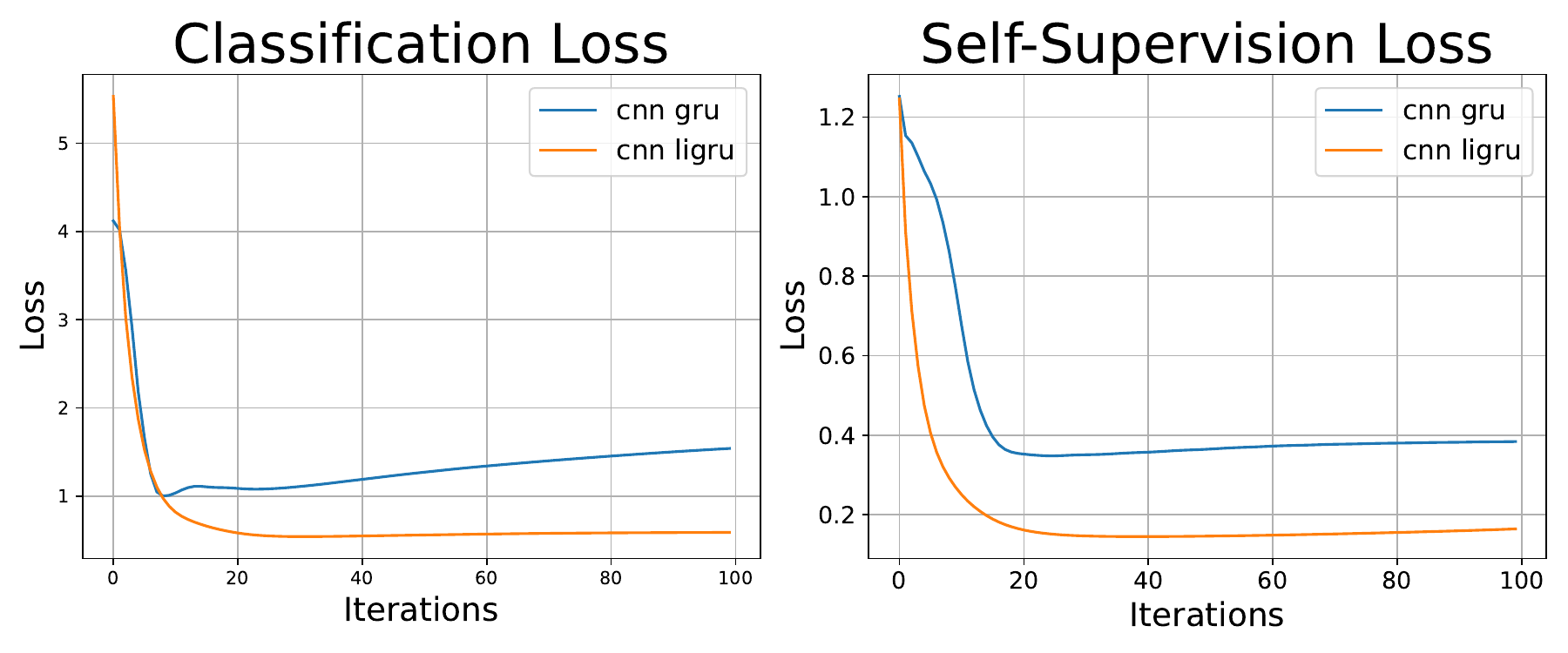}
        \caption{The loss value across iterations}
        \label{fig:visualize loss}
    \end{subfigure}
    \caption{\textbf{Left:} The change in norm of feature maps of Conv-GRU and Conv-LiGRU. \textbf{Right:} Loss value across iterations of Conv-GRU and Conv-LiGRU}
    \label{fig:layer_norm}
\end{figure}

However, we observe that this assumption is not entirely accurate. Figure \ref{fig:visualize norm} illustrates that $\|h_t - h_{t-1}\|_2$ of Conv-GRU converges to $0$ after more than $20$ iterations. 
Nevertheless, Figure \ref{fig:visualize loss} reveals that both the classification loss and self-supervised loss of the model exhibit divergence, corresponding to Conv-GRU encountering "overthinking" on corruption test sets, as shown in Figure \ref{fig:cnn_gru_gn_alpha_0.0}. 

In contrast, Conv-LiGRU demonstrates greater stability. Specifically, not only does $\|h_t - h_{t-1}\|_2$ converge to $0$ (Figure \ref{fig:visualize norm}), but both the main loss and auxiliary loss also converge smoothly to $0$ (Figure \ref{fig:visualize loss}). Additionally, Figure \ref{fig:cnn_ligru_alpha_0.0} shows that Conv-LiGRU significantly mitigates the "overthinking" phenomenon compared to Conv-GRU. 

In conclusion, the convergence of $\|h_t - h_{t-1}\|_2$ does not ensure that the model is free from "overthinking."

%% file: section/conclusion.tex
\section{Conclusion}
This study highlights the efficiency and adaptability of recurrent models for object recognition tasks. We demonstrate their strong generalization with fewer parameters compared to feedforward networks. To address the challenge of selecting optimal iterations during testing, we propose a self-supervised method to estimate accuracy trends, enhancing extrapolation capabilities. Additionally, we introduce Conv-LiGRU, a stable and efficient model that mitigates the "overthinking" issue and achieves superior accuracy, making it a robust choice for vision-based tasks. These findings pave the way for further advancements in lightweight and adaptive architectures for real-world applications.

\newpage

%% file: section/supplementary.tex
\onecolumn
\title{Learning to Stop Overthinking at Test Time\\(Supplementary Material)}
\maketitle

\appendix

\section{Background: Adaptive Computation Time (ACT)}
\label{sec:background}
ACT \cite{Graves2016AdaptiveComutationTime} is a mechanism designed to dynamically determine the number of recurrent steps required to process each input. Unlike its original formulation, which handles variable-length sequences, our work applies ACT to static inputs in visual reasoning tasks.  

At each time step, the model generates a halting score $p_t$ through a learned convolutional layer. The cumulative sum of these scores, $P_t$, determines whether the computation should continue. When $P_t$ reaches a predefined threshold $(1 - \epsilon)$, iteration stops, and the final hidden state is computed as a weighted sum of the intermediate states.  

A key component of ACT is the ponder cost, an auxiliary loss term that encourages the model to minimize the number of recurrent steps while maintaining accuracy. The total loss function consists of the task loss $L_{task}(y, \hat{y}_{act})$ and the ponder cost, weighted by a hyperparameter $\tau$ (Equal \ref{eq:act_loss}). By tuning $\tau$, we control the trade-off between computational efficiency and performance. In our study, we analyze the limitations of ACT’s early stopping heuristic and propose a self-supervised approach to better estimate the optimal number of iterations.  
\begin{equation}
    \mathcal{L} = \sum_{i=0}^{|\mathcal{D}|} \frac{1}{|\mathcal{D}|} L_{task}(y^i , \hat{y}^i_{\text{act}}) - \tau \sum_{t=1}^{t^i_{\text{halt}} - 1} p^i_t
    \label{eq:act_loss}
\end{equation}

\section{Additional simulation results}
\label{sec:addition_result}
\begin{figure*}[htbp]
    \centering
    \includegraphics[width=\linewidth]{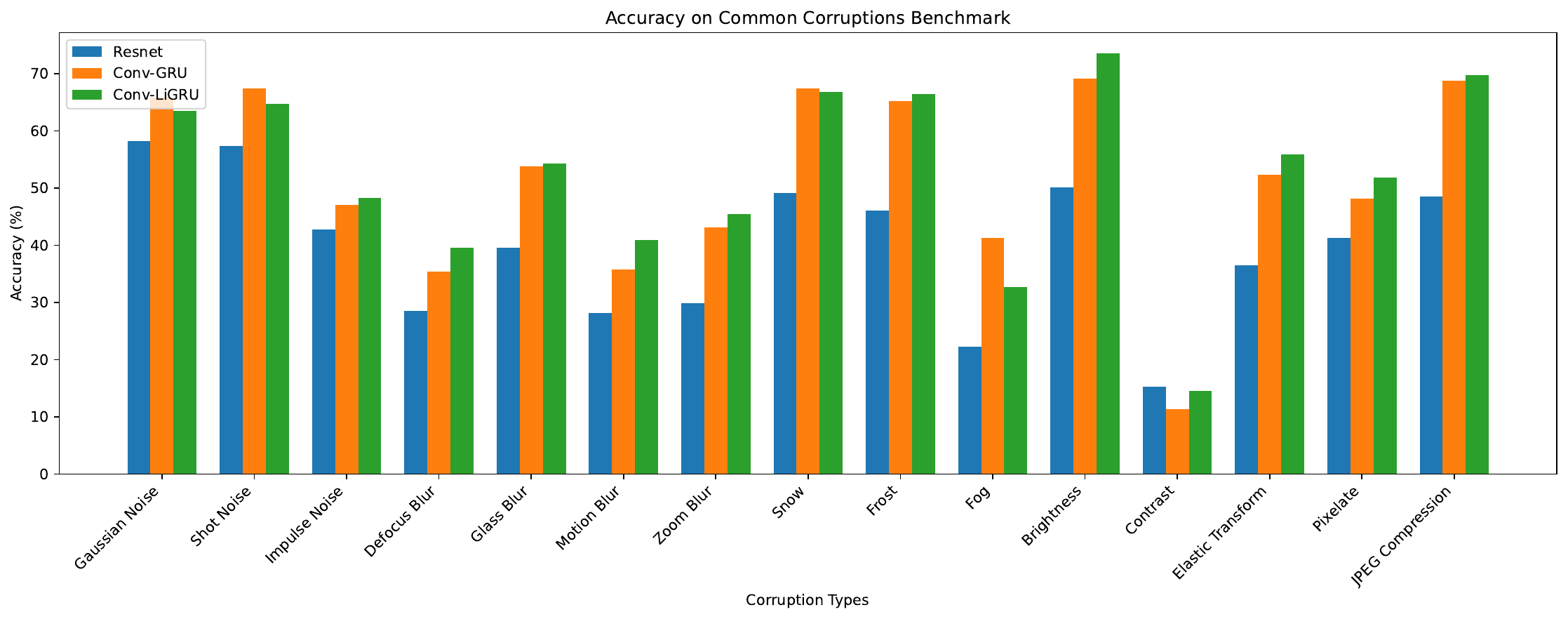}
        \caption{Accuracy (\%) on CIFAR10-C, at level 5, Resnet, Conv-GRU, Conv-LiGRU}
        \label{fig:bar_chart_cifar10c}
\end{figure*}

\begin{figure*}[htbp]
    \centering
    \includegraphics[width=\linewidth]{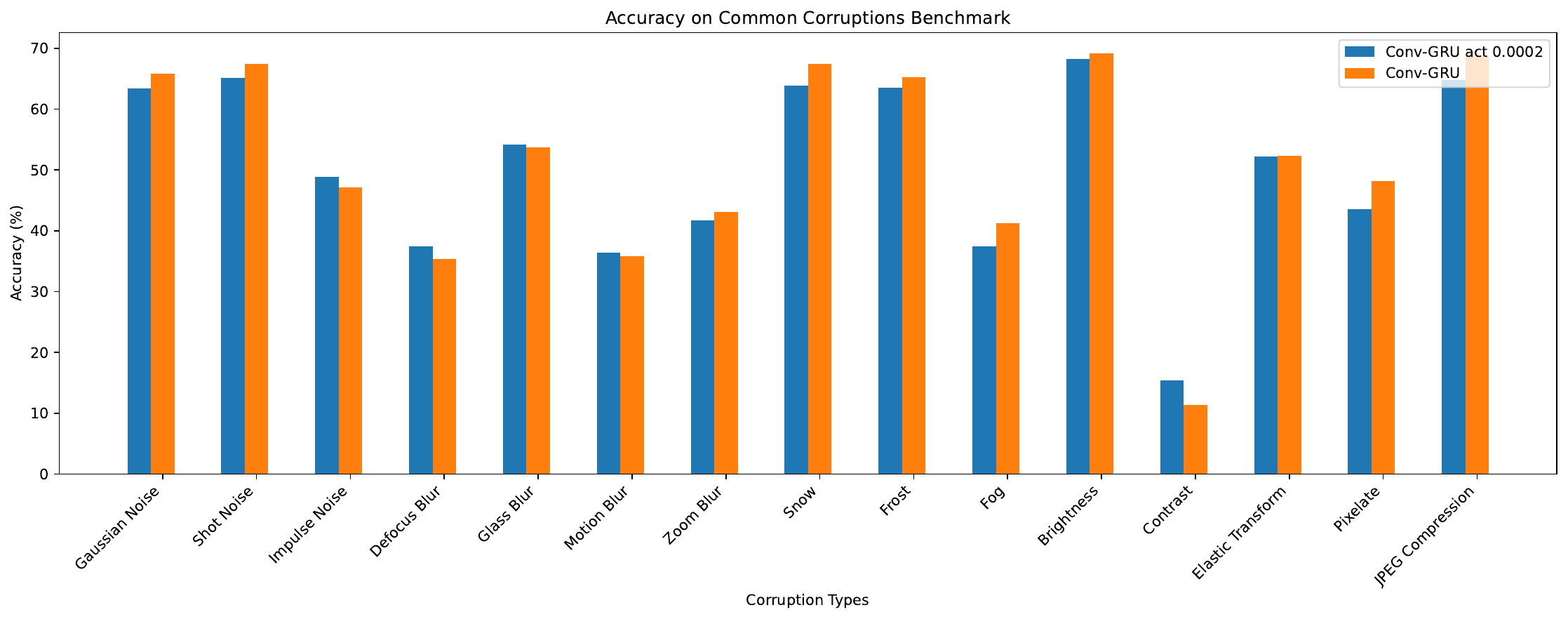}
        \caption{Accuracy (\%) on CIFAR10-C, at level 5, Conv-GRU with and without ACT}
        \label{fig:bar_chart_act}
\end{figure*}

\begin{table*}[t]
\centering
\caption{Test accuracy (\%) on CIFAR10-C, level 5}
\begin{tabular}{l|c|c|c|c|c|c|c}
\toprule
Corruptions & resnet & recall & cnn gru act 0.5 & cnn gru act 0.0002 & cnn gru & cnn ligru act 0.0002 & cnn ligru \\
\midrule
Gauss & 58.24 & 53.03 & 59.86 & 63.44 & 65.80 & 59.85 & 63.50 \\
Shot & 57.42 & 55.32 & 60.88 & 65.13 & 67.45 & 61.61 & 64.71 \\
Impul & 42.81 & 23.80 & 42.43 & 48.91 & 47.10 & 41.97 & 48.25 \\
Defoc & 28.55 & 28.83 & 37.96 & 37.50 & 35.36 & 41.17 & 39.61 \\
Glass & 39.53 & 42.78 & 53.30 & 54.21 & 53.78 & 50.24 & 54.26 \\
Motn & 28.19 & 28.14 & 38.70 & 36.36 & 35.81 & 40.97 & 40.97 \\
Zoom & 29.82 & 34.34 & 42.70 & 41.66 & 43.08 & 46.72 & 45.47 \\
Snow & 49.12 & 58.34 & 63.52 & 63.92 & 67.45 & 64.99 & 66.86 \\
Frost & 46.09 & 51.84 & 59.56 & 63.48 & 65.22 & 62.10 & 66.47 \\
Fog & 22.23 & 28.40 & 35.42 & 37.43 & 41.28 & 32.54 & 32.66 \\
Brit & 50.13 & 65.76 & 66.06 & 68.28 & 69.14 & 69.81 & 73.51 \\
Contr & 15.28 & 15.72 & 18.86 & 15.45 & 11.32 & 15.91 & 14.55 \\
Elast & 36.54 & 47.78 & 52.63 & 52.20 & 52.31 & 53.39 & 55.86 \\
Pixel & 41.27 & 28.89 & 51.30 & 43.57 & 48.16 & 45.68 & 51.82 \\
JPEG & 48.50 & 61.05 & 65.12 & 64.76 & 68.81 & 68.37 & 69.73 \\
\bottomrule
\end{tabular}
\label{tab:acc_all}
\end{table*}

\begin{table*}[t]
\setlength{\tabcolsep}{3pt} 
\centering
\caption{Test accuracy (\%) on CIFAR100-C, level 5, ResNet, Conv-GRU, Conv-LiGRU}
\begin{tabular}{l|ccccccccccccccc}
\toprule
& gauss & shot & impul & defoc & glass & motn & zoom & snow & frost & fog & brit & contr & elast & pixel & jpeg \\
\midrule
rn & 22.52 & 23.01 & 9.74 & 7.47 & 12.38 & 7.87 & 8.92 & 24.97 & 12.13 & 4.24 & 27.18 & 2.49 & 11.29 & 11.36 & 14.13 \\
cg & 26.02 & 25.85 & 10.73 & 16.64 & 23.00 & 18.70 & 20.00 & 35.26 & 32.01 & 10.88 & 39.61 & 4.41 & 31.28 & 19.74 & 34.95 \\
clig & 29.16 & 30.48 & 10.26 & 17.69 & 23.44 & 19.97 & 20.70 & 34.92 & 32.85 & 10.39 & 36.97 & 4.56 & 29.63 & 23.07 & 37.23 \\
\bottomrule
\end{tabular}
\label{tab:acc_cifar100}
\end{table*}

Figure \ref{fig:bar_chart_cifar10c} compares the performance of feedforward (ResNet) and recurrent architectures (Conv-GRU, Conv-LiGRU) in deep thinking networks on 15 corruption types at level 5 from the CIFAR10-C test set. The results show that recurrent models outperform feedforward networks in 14 out of 15 corruption types. Detailed accuracy values are provided in Table \ref{tab:acc_all}.

Additionally, we compare feedforward and recurrent models on CIFAR100-C, with Table \ref{tab:acc_cifar100} listing the accuracy of ResNet, Conv-GRU, and Conv-LiGRU on 15 corruption types at level 5. Both tables confirm the superiority of recurrent models over feedforward networks. Furthermore, Conv-LiGRU surpasses Conv-GRU on most test sets (11 out of 15 on CIFAR10-C and 12 out of 15 on CIFAR100-C), demonstrating its effectiveness and suitability for deep thinking models.

Figure \ref{fig:bar_chart_act} compares the performance of Conv-LiGRU with and without ACT. The results show that without ACT, Conv-LiGRU outperforms the ACT variant on 9 out of 15 level 5 test sets of CIFAR10-C. Detailed accuracy values for each test set are provided in Table \ref{tab:acc_all}. These findings reinforce that deep thinking models can achieve better performance when allowed to think freely rather than being constrained.
